%% file: MICCAI2026-main-conference-paper-template.tex
\documentclass[runningheads]{llncs}
\usepackage{booktabs}
\usepackage{amsmath}
\usepackage{bm}
\usepackage{graphicx}
\usepackage{xcolor}
\usepackage{cleveref}
\usepackage{tikz}
\usepackage{amssymb}
\usepackage{cite}
\usepackage{adjustbox}
\usepackage{tabularx}
\usepackage{siunitx}
\usepackage{multirow}
\usepackage{xspace}
\usepackage{multirow}
\usepackage[normalem]{ulem}
\useunder{\uline}{\ul}{}
\usepackage[table]{xcolor}
\usepackage{colortbl}

\usepackage[T1]{fontenc}
\usepackage{graphicx,verbatim}
\begin{document}
\title{Learning to Optimize Radiotherapy Plans \\via Fluence Maps Diffusion Model Generation \\and LSTM-based Optimization}
\titlerunning{Learning to Optimize Diffusion-based Radiation Therapy Planning}
%

\author{Isabella Poles\inst{1,2}\thanks{Corresponding author: \email{isabella.poles@polimi.it}} \and Simon Arberet\inst{2} \and Riqiang Gao\inst{2} \and  Martin Kraus\inst{3} \and Marco D. Santambrogio\inst{1} \and Florin C. Ghesu\inst{3} \and Ali Kamen\inst{2} \and Dorin Comaniciu\inst{2}}  
\authorrunning{I. Poles et al.}
\institute{Politecnico di Milano, Milan, Italy \and Digital Technology and Innovation, Siemens Healthineers, Princeton, NJ, USA \and Digital Technology and Innovation, Siemens Healthineers, Erlangen, Germany} 
  
\maketitle              
\begin{abstract}
Volumetric Modulated Arc Therapy (VMAT) is a cornerstone of modern radiation therapy, enabling highly conformal tumor irradiation and healthy-tissue sparing. Yet, its planning solves inverse and nested optimization for multi-leaf collimators, monitor units and dose parameters, while enforcing their consistency to ensure mechanical deliverability. Nevertheless, this process often requires repeated re-optimization when treatment configurations change, resulting in substantial planning time per patient. To address these problems, we present a diffusion-driven Learning-to-Optimize (L2O) method for end-to-end VMAT planning. A distribution-matching distilled diffusion model learns a clinically feasible manifold of fluence maps, enabling their one-shot generation. On top of this, an LSTM-based L2O module learns gradient update dynamics to swiftly refine fluence maps toward prescribed dose objectives during inference. Experimental results on clinical and public prostate cancer cohorts demonstrate improved planning efficiency, flexibility, and machine deliverability over currently available end-to-end VMAT planners.


\end{abstract}
\section{Introduction}\label{sec:introduction}
\input{sections/intro}

\section{Methodology}\label{sec:impl}
\input{sections/implementation}

\section{Experiments}\label{sec:res}
\input{sections/results}

\section{Conclusion}\label{sec:conclusions}
\input{sections/conclusions}

\begin{credits}
\subsubsection{\disclaimername} The concepts and information presented in this paper are based on research results that are not commercially available. Future availability cannot be guaranteed.

\subsubsection{\ackname} All contributors to the REQUITE project are acknowledged, including the patients, clinicians, and nurses. The consortium consists of David Azria, Erik Briers, Jenny Chang-Claude, Alison M. Dunning, Rebecca M. Elliott, Corinne Faivre-Finn, Sara Gutierrez-Enriquez, Kerstie Johnson, Zoe Lingard, Tiziana Rancati, Tim Rattay, Barry S. Rosenstein, Dirk De Ruysscher, Petra Seibold, Elena Sperk, R. Paul Symonds, Hilary Stobart, Christopher Talbot, Ana Vega, Liv Veldeman, Tim Ward, Adam Webb and Catharine M.L. West.
\end{credits}


%
%
%
\bibliographystyle{splncs04}
\bibliography{reference}






\end{document}

%% file: sections/intro.tex
Radiation Therapy (RT) is a core component of modern cancer treatment, enabling irradiation of tumors to impair cell proliferation and halt disease progression~\cite{jaffray2023harnessing}. Intensity Modulated RT (IMRT) and Volumetric Modulated Arc Therapy (VMAT) are the most widely adopted techniques~\cite{liu2018comparison}. While both are effective, VMAT enhances target dose conformity and healthy‑tissue sparing via fast, continuous radiation modulation as the machine's Multi-Leaf Collimator (MLC) rotates around the patient.
Despite this, its planning is a high-dimensional inverse problem with nested optimization of the MLC radiation aperture, Monitor Units (MU) intensity, and dose objectives, which are adjusted until a clinically acceptable plan is achieved. Moreover, machine-dose \textit{consistency} must be enforced to ensure that the planned dose remains deliverable under machine constraints, further complicating the process. Consequently, current pipelines often require tens of minutes per patient and per treatment-configuration change~\cite{claessens2022quality}.

Motivated by the need for efficient, flexible, and dose-machine-consistent solutions, we shift the VMAT planning paradigm from a conventional optimization to an end-to-end diffusion-based Learning-to-Optimize (L2O) formulation. 

Some recent end‑to‑end VMAT planners optimize MLC apertures and MU via gradient‑based Direct Aperture Optimization (DAO)~\cite{simko2025physics} and (un)constrained optimizers~\cite{kingma2014adam, liu2021neuraldao, dursun2023automated, dubois2023radiotherapy, zhang2025aperture}, allowing DL‑based dose predictors~\cite{gao2023flexible,feng2025leveraging,gao2025generative} and engine simulation~\cite{kraus2026single} to be embedded in differentiable pipelines, potentially accelerating planning. Other approaches cast planning as a decision-making problem using Reinforcement Learning (RL)~\cite{hrinivich2020artificial,kafaei2021graph}, 
or as a supervised inference task enabling single forward-pass plan prediction~\cite{yang2025foresight, arberet2026ai}. %
Nevertheless, the literature highlights three key limitations.
First, certain DAO planners report piecewise‑differentiable and plateau‑heavy loss landscapes due to MLC quantization and gap constraints, leading to illusory/slow convergence, which hinders practical deployment~\cite{wu2025machine}.
Second, some RL‑based approaches adjust machine parameters through discrete or partially continuous actions that inadequately explore the full MLC search space~\cite{hrinivich2020artificial}. Convolutional and multi‑agent policies solve the issue, enabling continuous dose‑rate and MLC modulation, but requiring large RT‑plan datasets for generalization~\cite{mekki2025dual} and remaining restricted to leaf‑sequencing tasks based on pre‑optimized fluence maps beam‑intensity patterns~\cite{gao2024multi}.
Third, supervised approaches can be hindered by the non‑uniqueness of machine‑parameter-dose mappings and limited adaptability to changing planning conditions~\cite{heilemann2025ultra}.

To address these gaps, we propose a unified fluence‑map generation L2O method achieving fast, flexible, and clinically feasible end-to-end VMAT planning, as shown in \Cref{fig:meth}. 
Indeed, fluence maps provide a differentiable parameterization of the beam‑intensity modulation achievable through feasible MLC motion and MU patterns, yielding a smooth representation that supports stable gradient-based updates while promoting physically deliverable patterns~\cite{arberet2025beam}.
Therefore, our primary goal is to update fluence maps iteratively, leveraging the \textit{smooth} gradient landscape computed between a DL‑predicted and clinically desired dose to satisfy dose-machine-consistent clinical objectives and to \textit{learn} efficient \textit{optimization} dynamics.
To achieve this,
we employ diffusion models to model the non-unique distribution of clinically feasible planning solutions, and leverage \textbf{distribution-matching distillation for efficient single-shot fluence map generation}. 
Operating in this manifold without relying on target fluence map initializations, the LSTM‑based \textbf{L2O module} learns the history of the optimization gradients during training, \textbf{finding} \textbf{the fast and saddle-points free pathway to plan solution} during inference. 
Unlike existing methods, \textbf{our approach is initialization agnostic}, \textbf{efficient}, \textbf{flexible} to changing objectives \textbf{without retraining} and \textbf{guarantees plan delivery}. 

We summarize our contributions as follows: (1) \textsc{FMD}: a one-shot \underline{F}luence \underline{M}ap \underline{D}iffusion model trained via distribution-matching distillation to generate VMAT fluence maps and a manifold of non‑unique plan solutions; (2) \textsc{L2Plan}: a novel LSTM-based \underline{L2}O VMAT \underline{Plan}s module that learns iterative update dynamics to efficiently refine fluence maps toward prescribed dose objectives without tuned initializations or retraining; (3) A validation on three prostate cancer clinical and public patient cohorts demonstrating improved efficiency, flexibility to changing dose objectives, and plan deliverability verified via machine leaf sequencing compared to existing optimization and learning-based approaches.

%% file: sections/implementation.tex
\begin{figure}[t]
\centering
\includegraphics[width=.95\textwidth]{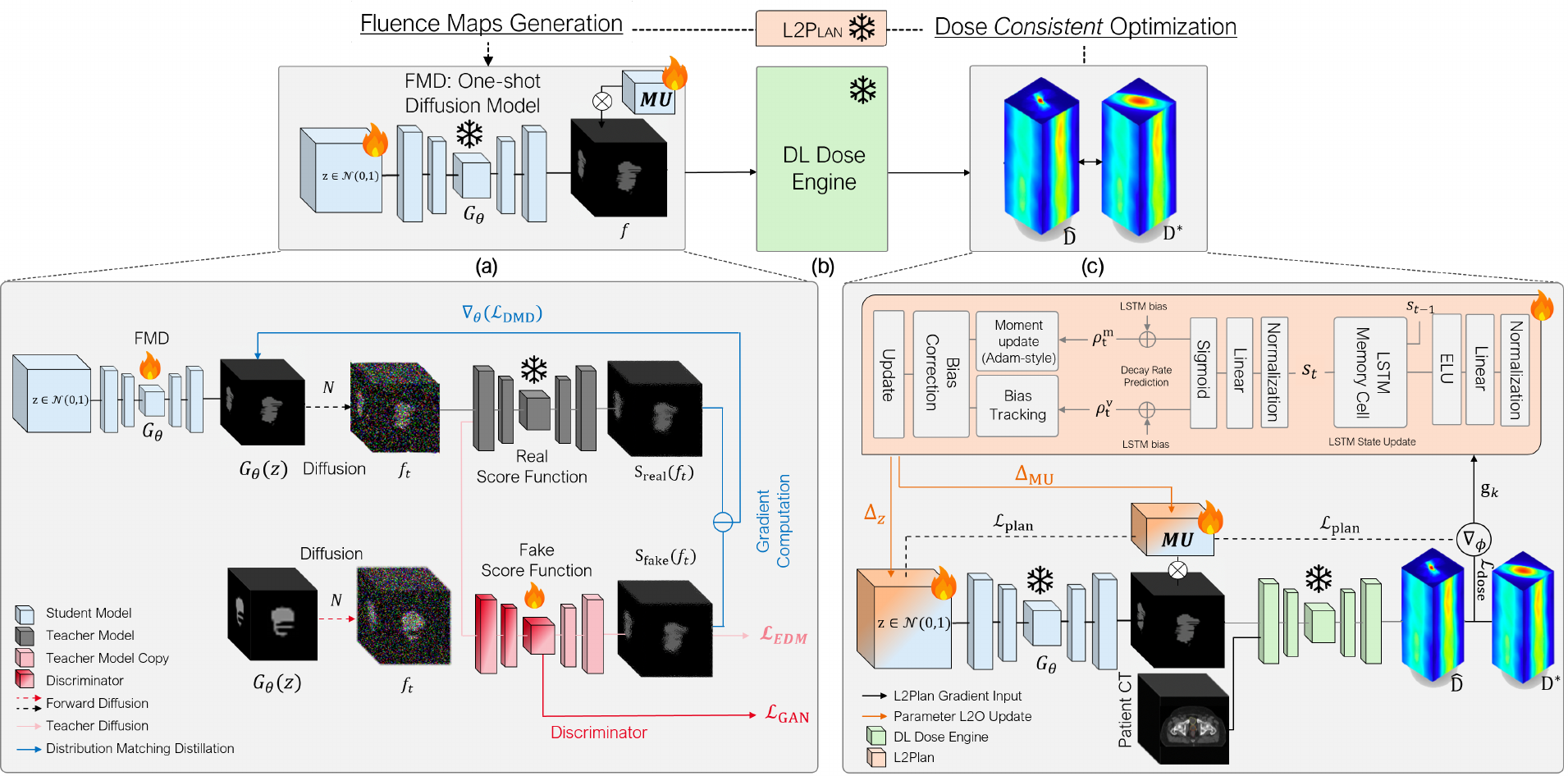}
\caption{Overview of our method from the \textsc{FMD} model, which generates fluence maps in one-shot (a), to the \textsc{L2Plan} optimizer (c), which learns to optimize fluence maps so that the corresponding predicted dose (b) matches the target plan.} \label{fig:meth}
\end{figure}

Our methodology combines diffusion-based fluence generation with L2O refinement to enforce dose-machine consistency, ensuring a prescribed dose $\mathbf{D}^\star$ admits at least one feasible fluence map $\bm{f}\in \mathcal{F}_{\text{del}}$, where $\mathcal{F}_{\text{del}}$ denotes the set of machine-constrained fluence maps defined by limits on leaf motion, dose rate, and gantry rotation, and validated via MLC Leaf Sequencing (LS).

First, we distill a diffusion model trained on $\mathcal{F}_{\text{del}}$ into a one-shot generator $G_\theta$ that rapidly samples a plausible initial fluence $\bm{f} = G_\theta(\mathbf{z})$ from latent noise $\mathbf{z} \sim \mathcal{N}(0,I)$, without iterative denoising. Second, we train an LSTM-based optimizer to update $\mathbf{z}$ and a $\mathbf{MU}$ parameter to scale $\bm{f}$ intensity, $\bm{\phi}=\{\mathbf{z},\mathbf{MU}\}$, using the dose gradients history between the planned dose $\mathbf{D}^\star$ and the predicted one $\hat{\mathbf{D}}$, obtained from $\bm{f}$ and a patient Computed Tomography (CT) image via a differentiable dose engine~\cite{kraus2026single}. The diffusion and dose modules are trained separately and kept frozen during LSTM training; all modules remain frozen at inference, enabling rapid refinement of $\bm{f}$ via $\bm{\phi}$ while preserving deliverability. An overview is provided in \Cref{fig:meth}.

\subsection{\textsc{FMD}: Fluence Maps Diffusion Model Generation in One-shot}
Despite their ability to model high-quality deliverable VMAT fluence maps, diffusion models require expensive iterative denoising, making distillation into a one-shot generator imperative for efficient planning. Therefore, to guarantee high-quality generation and facilitate the subsequent distillation process, we first train a Dhariwal-style UNet as the teacher diffusion model, employing four downsampling stages with residual blocks, group normalization, SiLU activations, and a symmetric skip-connected decoder following EDM~\cite{karras2022elucidating}. In particular, in the forward diffusion process, a fluence map $\bm{f} \sim \mathcal{F}_{\text{real}}$ is progressively corrupted by injecting Gaussian noise across $T$ continuous noise levels, such that at noise level $t$ it follows the marginal distribution
$\mathcal{F}_{\mathrm{real},t}(f_t)=\int \mathcal{F}_{\mathrm{real}}(\bm{f})\,z_t(f_t\mid \bm{f})\,df$,
where $z_t(f_t\mid \bm{f})=\mathcal{N}(\alpha_t \bm{f},\sigma_t^2 \bm{I})$ and $\alpha_t,\sigma_t>0$ are defined by the EDM noise schedule.
Teacher reverse diffusion is learned via denoising score matching by minimizing:
\begin{equation}
\mathcal{L}_{\mathrm{EDM}}
=
\mathbb{E}_{\bm{f}\sim \mathcal{F}_{\mathrm{real}},\,t\sim[0,T]}
\bigl\|
\mu_{\mathrm{real}}(f_t,t)-\bm{f}
\bigr\|_2^2,
\label{eq:edm_loss}
\end{equation}
which estimates a clean fluence map $\mu_{\mathrm{real}}$ from noisy ones and, implicitly, the score function
$\bm{s}_{\mathrm{real}}(f_t,t)=\nabla_{f_t}\log \mathcal{F}_{\mathrm{real},t}(f_t)$ of the diffused distribution $\mathcal{F}_{\mathrm{real}}$.

To enable one-shot sampling, we distill the teacher into a single-step generator $G_\theta$ via Distribution Matching Distillation (DMD)~\cite{yin2024improved}, minimizing the expected Kullback-Leibler (KL) divergence between the teacher $\mathcal{F}_{\mathrm{real},t}$ and student  $\mathcal{F}_{\mathrm{fake},t}$ diffused data distributions.
Since $G_\theta$ is trained by minimizing KL loss between the teacher-student diffusion distributions, whose likelihood is intractable, we directly compute the gradient of the KL loss, which depends just on the difference between their score functions, making explicit density evaluation unnecessary:
\begin{equation}
\nabla_\theta \mathcal{L}_{\mathrm{DMD}}
=
-\mathbb{E}_{z\sim\mathcal{N}(0,I),\,t}
\Bigl[
\bigl(
\bm{s}_{\mathrm{real}}(f_t,t)-\bm{s}_{\mathrm{fake}}(f_t,t)
\bigr)
\,
\frac{\partial G_\theta(z)}{\partial\theta}
\Bigr],
\label{eq:dmd_grad}
\end{equation}
where $f_t = N(G_\theta(z),t)$, $N$ denotes the forward noising process, and $s_{\mathrm{fake}}$ is estimated by a teacher copy trained online on the $G_\theta$ generated samples.
Since the $G_\theta$ never observes real data, inheriting approximation errors from the teacher, we anchor $G_\theta$ to real data, by training a discriminator $\mathcal{D}_\text{GAN}$ on noisy samples~\cite{yin2024improved}:
\begin{equation}
\mathcal{L}_{\mathrm{GAN}}
=
\mathbb{E}_{\bm{f}\sim \mathcal{F}_{\mathrm{real}},\,t}
\bigl[\log \mathcal{D}_{\text{GAN}}(N(f_{t},t))\bigr]
+
\mathbb{E}_{\bm{z}\sim\mathcal{N}(0,I),\,t}
\bigl[\log(1-\mathcal{D}_{\text{GAN}}(N(G_\theta(z),t)))\bigr],
\label{eq:gan_loss}
\end{equation}
which gradient jointly with $\nabla_\theta \mathcal{L}_{\mathrm{DMD}}$ update $G_\theta$.
The resulting updates are independent of the teacher's sampling trajectories and mitigate score approximation errors, allowing $G_\theta$ to match the teacher's sample quality in a single step.

\subsection{\textsc{L2Plan}: Learning-to-Optimize Dose Consistent RT Plans}
After generating an initial $\bm{f}$, we enforce consistency by $\bm{\phi}=\{\mathbf{z},\mathbf{MU}\}$ updating via backpropagation of the plan loss $\mathcal{L}_{\text{plan}}(\hat{\mathbf{D}},\mathbf{D}^{\star})$, driving the predicted dose $\hat{\mathbf{D}}$ toward the target one $\mathbf{D}^{\star}$. Although $\bm{\phi}$ end-to-end optimization is feasible, standard gradient-based methods suffer from instability/saddle-point stagnation, as even simple dose objectives induce a non-convex landscape when composed with the nonlinear fluence-to-dose mapping, which static hyperparameters struggle to adapt to. As a result, optimization may drift toward numerically optimal yet physically inconsistent solutions, characterized by needle-like apertures or spiky MU profiles that violate machine constraints.

To address this issue, we adopt an L2O strategy that jointly trains a recurrent optimizer to learn task-adaptive update rules that escape saddle points and enable fast dose-driven parameter refinement. \textsc{L2Plan} consists of an \textit{inner optimizee loop}, which iteratively updates $\bm{\phi}$ through learned hyperparameterized update rules, and an \textit{outer meta-optimization loop}, which learns the parameters of the meta-optimizer itself. Starting from randomly initialized meta-optimizer parameters at iteration $k=0$, the outer loop performs $K_0$ meta-optimization steps, where each outer update is driven by the outcome of an inner optimizee unrolled for $k_i$ steps. In the inner loop, at step $k$, the dose objective gradient $\mathbf{g}_k=\nabla_{\phi_k}\mathcal{L}_{\text{plan}}$ is normalized, passed through a learnable linear preprocessing function $\delta(\cdot)$ and used to update a latent optimizer state $\mathbf{s}_k$ which implicitly track the gradient history via an LSTM with two recurrent layers: 
\begin{equation}
\mathbf{s}_k=\mathrm{LSTM}(\delta(\mathbf{g}_k),\mathbf{s}_{k-1}).
\end{equation}
Conditioned on the current state $\mathbf{s}_k$, the meta-optimizer predicts hyperparameters inspired by adaptive first-order optimization methods, namely the first and second order momentum decay rates $\beta_k$ and $\gamma_k$, through meta-learnable linear projections $l(\cdot)$ followed by $\sigma$ sigmoid activations to ensure valid ranges: 
\begin{equation}
\beta_k=\sigma(l_{\beta}(\mathbf{s}_k)), \quad \gamma_k=\sigma(l_{\gamma}(\mathbf{s}_k)).
\end{equation}
These parameters are initialized to standard Adam values at $k=0$ and subsequently adapted according to the gradient signals produced during the learned optimization process. Using the $\beta_k$, $\gamma_k$ and $\mathbf{g}_k$, moment estimates are updated as $\mathbf{m}_k=\beta_k\mathbf{m}_{k-1}+(1-\beta_k)\mathbf{g}_k$ and $\mathbf{v}_k=\gamma_k\mathbf{v}_{k-1}+(1-\gamma_k)\mathbf{g}_k^2$, with bias correction applied. Given the predicted update direction, the resulting optimizee step $\Delta\phi_k$ and parameter update $\phi_{k+1}$ are computed as:
\begin{equation}
\Delta\phi_k=-\hat{\mathbf{m}}_k/(\sqrt{\hat{\mathbf{v}}_k}+\epsilon) \rightarrow \phi_{k+1}=\phi_k+\Delta\phi_k.
\end{equation}
After $k_i$ inner‑loop updates are performed, $\mathcal{L}_{\text{plan}}$ updates the outer LSTM parameters. The trained meta‑optimizer is then frozen, guaranteeing the $\mathbf{D}^{\star}$ clinical objective with the fewest $k$ prediction steps $\Delta\phi_k$ during planning.

\subsection{Overall Framework}
In our approach, we formulate VMAT planning as a Maximum A Posteriori (MAP) problem which estimates $\bm{\phi}=\{\mathbf{z},\mathbf{MU}\}$ values most consistent with the observed $\mathbf{D}^{\star}$ while respecting some prior assumptions. The MAP objective $\mathcal{L}_{\text{plan}}
=
\mathcal{L}_{\text{dose}}
+
\lambda_{\mathbf{z}}\mathcal{L}_{\text{cont}}^{\mathbf{z}}
+
\lambda_{\mathbf{MU}}\mathcal{L}_{\text{cont}}^{\mathbf{MU}}$ is minimized with the learned meta‑optimizer over 100 steps $k$, consistently achieving convergence.
Here, the likelihood term $\mathcal{L}_{\text{dose}} = \|\hat{\mathbf{D}}-\mathbf{D}^{\star}\|_{1}$ enforces fidelity to the target dose, while the continuity priors $\mathcal{L}_{\text{cont}}^{\mathbf{z}}
=
\sum_{\text{CP}}\|\mathbf{z}_{\text{CP}+1}-\mathbf{z}_{\text{CP}}\|_1$ and
$\mathcal{L}_{\text{cont}}^{\mathbf{MU}}
=
\sum_{\text{CP}}\|\mathbf{MU}_{\text{CP}+1}-\mathbf{MU}_{\text{CP}}\|_1$ encourage smooth transitions across Control Points (CPs). We fix the regularization weights to $\lambda_{\mathbf{z}}=5$ and $\lambda_{\mathbf{MU}}=1$ throughout our experiments.

%% file: sections/results.tex
\subsection{Experimental Setup} 
\subsubsection{Dataset.} We curated a large cohort from the publicly available REQUITE dataset~\cite{seibold2019requite} comprising 12,469 single‑arc prostate VMAT plans and targeting the whole prostate, seminal vesicles, post‑surgical prostate bed, and pelvic lymph‑node PTVs, while sparing bladder and rectum OARs. 
Across the cohort, plans have fixed $\ang{30}$ collimator and $\ang{0}$ couch angles. Each clockwise arc spans nearly a full rotation and is discretized into 178 CPs. All treatments employed the Varian Millennium 120 (M120) MLC with 60 leaf pairs, 5$mm$-10$mm$ wide, in the central 40 and the outer 20 pairs, respectively.
\begin{table}[t!]
\centering
\caption{Results on the REQUITE dataset. \lq\lq Optimizer\rq\rq\ refers to the strategy enforcing dose-machine consistency; \lq\lq Generator\rq\rq\ denotes the fluence‑map model; \lq\lq Steps\rq\rq\ the number of optimization iterations; and \lq\lq LS\rq\rq\ rule-based Leaf Sequencing results. The best results are in bold, while (*) indicates 0.001 p-value significant difference.}
\label{tab::res}
\centering
\begin{adjustbox}{width=\textwidth}
\begin{tabular}{@{}lllccccccccc@{}}
\toprule
\multicolumn{2}{c}{\textbf{Optimizer}} & \multicolumn{1}{c}{\textbf{Generator}} & \textbf{Steps} & \multicolumn{1}{c}{\textbf{FID} $\downarrow$} &  \multicolumn{1}{c}{\textbf{MSE} $\downarrow$} & \multicolumn{1}{c}{\textbf{SSIM} $\uparrow$} & \multicolumn{1}{c}{\textbf{PSNR} $\uparrow$} & \multicolumn{1}{c}{\textbf{MAE} $\downarrow$} & \multicolumn{1}{c}{\textbf{MAE$_{\text{PTV}}$} $\downarrow$} & \multicolumn{1}{c}{\textbf{MAE$_{\text{OARs}}$} $\downarrow$} & \multicolumn{1}{c}{\textbf{Time} $\downarrow$} \\ \midrule
\multicolumn{10}{l}{\textit{Standard VMAT Optimizers}} \\ \midrule
\multicolumn{2}{l}{RMSProp~\cite{dubois2023radiotherapy}} & \textsc{FMD} & 300 & \multicolumn{1}{c}{17.52} & \multicolumn{1}{c}{-} & 0.940$\pm$0.040* & 32.66$\pm$4.35* &  0.173$\pm$0.065* & 1.151$\pm$0.224* & 0.422$\pm$0.298* & 504.0$\pm$1.1* \\
\multicolumn{2}{l}{L-BFGS~\cite{wu2026illusion}} & \textsc{FMD} & 300 & \multicolumn{1}{c}{17.52} & \multicolumn{1}{c}{-} & 0.971$\pm$0.008* & 37.91$\pm$1.67* & 0.099$\pm$0.039*  & 0.638$\pm$0.198* & 0.255$\pm$0.087* & 890.4$\pm$3.5*\\
\multicolumn{2}{l}{SGD$_\text{M}$~\cite{dubois2023radiotherapy}} & \textsc{FMD} & 300 & \multicolumn{1}{c}{17.52} & \multicolumn{1}{c}{-} & 0.978$\pm$0.009* & 38.48$\pm$1.84* & 0.089$\pm$0.024*  & 0.431$\pm$0.292* & 0.183$\pm$0.092* & 508.2$\pm$1.6* \\
\multicolumn{2}{l}{Adam~\cite{kingma2014adam}} & \textsc{FMD} & 300 & \multicolumn{1}{c}{17.52} & \multicolumn{1}{c}{-} & 0.985$\pm$0.007* & 39.47$\pm$2.56* & 0.085$\pm$0.032* & 0.720$\pm$0.270* & 0.231$\pm$0.131* & 505.1$\pm$1.2* \\
\multicolumn{2}{l}{DAO~\cite{simko2025physics}} & - & 2000 & - & - & 0.919$\pm$0.071* & 28.02$\pm$4.28* & 0.324$\pm$0.070* & 1.061$\pm$0.545* & 0.747$\pm$0.071* & 2966.4$\pm$9.4* \\
\multicolumn{2}{l}{\textbf{Adam~\cite{kingma2014adam}}} & \textbf{\textsc{FMD}} & \textbf{2000} & \multicolumn{1}{c}{\textbf{17.52}} & \multicolumn{1}{c}{\textbf{-}} & \textbf{0.998$\pm$0.001} & \textbf{50.90$\pm$3.24} & \textbf{0.019$\pm$0.009} & \textbf{0.170$\pm$0.070}& \textbf{0.060$\pm$0.041} & \textbf{2339.2$\pm$7.2} \\ 
\midrule
\multicolumn{10}{l}{\textit{Generative Models}} \\ \midrule
\multicolumn{2}{l}{Adam~\cite{kingma2014adam}} & VQ-VAE~\cite{guo2025maisi} & 300 & \multicolumn{1}{c}{-} & \multicolumn{1}{c}{5e-3} & 0.988$\pm$0.004* & 37.71$\pm$1.49* & 0.080$\pm$0.028* & 1.221$\pm$0.329* & 0.395$\pm$0.194* & 547.6$\pm$1.5*\\
\multicolumn{2}{l}{Adam~\cite{kingma2014adam}} & StyleGAN2~\cite{karras2020analyzing} & 300 &  \multicolumn{1}{c}{32.94} & \multicolumn{1}{c}{-} &  0.979$\pm$0.006* & 38.40$\pm$1.51* & 0.113$\pm$0.044* & 0.920$\pm$0.190* & 0.440$\pm$0.280* & 469.0$\pm$2.0*\\
\multicolumn{2}{l}{Adam~\cite{kingma2014adam}} & D2O~\cite{zheng2025revisiting} & 300 & \multicolumn{1}{c}{23.76} & \multicolumn{1}{c}{-} & 0.963$\pm$0.125* & 38.06$\pm$6.46* & 0.098$\pm$0.111* &  0.963$\pm$0.397* & 0.511$\pm$0.623* & 487.2$\pm$1.9*\\\multicolumn{2}{l}{\textbf{Adam~\cite{kingma2014adam}}} & \textbf{\textsc{FMD}} & \textbf{300} & \multicolumn{1}{c}{\textbf{17.52}} & \multicolumn{1}{c}{-} & \textbf{0.985$\pm$0.007*} & \textbf{39.47$\pm$2.56*} & \textbf{0.085$\pm$0.032*} & \textbf{0.720$\pm$0.270*} & \textbf{0.231$\pm$0.131*} & \textbf{505.1$\pm$1.2*} \\ \midrule
\multicolumn{10}{l}{\textit{L2O Optimizers}} \\ \midrule
\multicolumn{2}{l}{VeLO~\cite{janson2025pylo}} & \textsc{FMD} & 300 & \multicolumn{1}{c}{17.52} & \multicolumn{1}{c}{-} & 0.985$\pm$0.015* & 40.62$\pm$3.75* & 0.066$\pm$0.018* & 0.471$\pm$0.033* & 0.243$\pm$0.120* & 461.1$\pm$3.1*
  \\
\multicolumn{2}{l}{$\mu$LO~\cite{therien2024mu}} & \textsc{FMD} & 300 & \multicolumn{1}{c}{17.52} & \multicolumn{1}{c}{-} & 0.972$\pm$0.030* & 40.54$\pm$0.95* & 0.091$\pm$0.021* & 0.222$\pm$0.040* & 0.221$\pm$0.193*  & 458.3$\pm$1.5*   \\
\multicolumn{2}{l}{CoordMath~\cite{liu2023towards}} & \textsc{FMD} & 300 & \multicolumn{1}{c}{17.52} & \multicolumn{1}{c}{-} & 0.981$\pm$0.022* & 39.66$\pm$1.09* & 0.097$\pm$0.017* & 0.345$\pm$0.048* & 0.299$\pm$0.201*  & 467.1$\pm$1.3*  \\
\multicolumn{2}{l}{HyperAdam~\cite{wang2019hyperadam}} & \textsc{FMD} & 300 & \multicolumn{1}{c}{17.52} & \multicolumn{1}{c}{-} & 0.990$\pm$0.020* & 41.72$\pm$1.11* & 0.054$\pm$0.011* & 0.207$\pm$0.090* & 0.255$\pm$0.107*  & 450.3$\pm$1.0*    \\
\multicolumn{2}{l}{\textbf{\textsc{L2Plan}}} & \textbf{\textsc{FMD}} & \textbf{300} & \multicolumn{1}{c}{\textbf{17.52}} & \multicolumn{1}{c}{-} & \textbf{0.998$\pm$0.001} &  \textbf{50.61$\pm$1.84} & \textbf{0.026$\pm$0.009}  & \textbf{0.149$\pm$0.076} &  \textbf{0.061$\pm$0.039} & \textbf{478.0$\pm$0.9} \\
\multicolumn{2}{l}{\cellcolor[rgb]{0.89,0.89,0.89}\textbf{\textsc{L2Plan}}} & \cellcolor[rgb]{0.89,0.89,0.89}\textbf{\textsc{FMD}} & \cellcolor[rgb]{0.89,0.89,0.89}\textbf{100} & \multicolumn{1}{c}{\cellcolor[rgb]{0.89,0.89,0.89}\textbf{17.52}} & \multicolumn{1}{>{\columncolor[rgb]{0.89,0.89,0.89}}c}{\textbf{-}} & \cellcolor[rgb]{0.89,0.89,0.89}\textbf{0.993$\pm$0.003} & \cellcolor[rgb]{0.89,0.89,0.89}\textbf{45.33$\pm$1.62} & \cellcolor[rgb]{0.89,0.89,0.89}\textbf{0.059$\pm$0.019}  & \cellcolor[rgb]{0.89,0.89,0.89}\textbf{0.445$\pm$0.008} &\cellcolor[rgb]{0.89,0.89,0.89}\textbf{0.154$\pm$0.087}& \cellcolor[rgb]{0.89,0.89,0.89}\textbf{159.9$\pm$0.5}  \\ 
\multicolumn{2}{l}{\cellcolor[rgb]{0.89,0.89,0.89}\textbf{\textsc{L2Plan}$_\text{LS}$}} & \cellcolor[rgb]{0.89,0.89,0.89}\textbf{\textsc{FMD}} & \cellcolor[rgb]{0.89,0.89,0.89}\textbf{100} & \multicolumn{1}{c}{\cellcolor[rgb]{0.89,0.89,0.89}\textbf{17.52}}  & \multicolumn{1}{>{\columncolor[rgb]{0.89,0.89,0.89}}c}{\textbf{-}} & \cellcolor[rgb]{0.89,0.89,0.89}\textbf{0.993$\pm$0.002}  & \cellcolor[rgb]{0.89,0.89,0.89}\textbf{45.22$\pm$1.53} & \cellcolor[rgb]{0.89,0.89,0.89}\textbf{0.058$\pm$0.019} & \cellcolor[rgb]{0.89,0.89,0.89}\textbf{0.445$\pm$0.098} & \cellcolor[rgb]{0.89,0.89,0.89}\textbf{0.163$\pm$0.084}& \cellcolor[rgb]{0.89,0.89,0.89}\textbf{161.0$\pm$0.4} \\\bottomrule
\end{tabular}
\end{adjustbox}
\end{table}
\subsubsection{Implementation Details.} 
Treatment plans undergo data augmentation by generating synthetic single-arc plan variants using Eclipse TPS with randomized field size, gantry angle variations, and 0.5$\times$-1.5$\times$ fluence‑map intensity scaling relative to the original value. Diffusion‑model and LSTM‑based meta‑optimizations are performed using Adam (learning rate 1$\cdot$10$^{-5}$, momentum 0.9) with batch sizes of 256 and 4, respectively, implemented in PyTorch (1.13.0) on 4$\times$NVIDIA H100 GPUs (80GB). The dataset is split per patient into 12,153-79-79 train-validation-test cases, and results are reported with paired $t$-test analysis.

\subsection{Comparison with Currently Available VMAT Planners}
We compared our method against RT, latent-variable generative, and L2O baselines at their convergence step. As shown in \Cref{tab::res}, \textsc{L2Plan} achieves higher accuracy with fewer optimization iterations than standard approaches.  
Specifically, it improves PTV coverage in Mean Absolute Error (MAE) (0.720 \textit{v.s.} 0.149 MAE$_{PTV}$) and OARs sparing (0.231 \textit{v.s.} 0.061 MAE$_{OARs}$) over unconstrained RT planners with the same 300 iterations budget, while reaching comparable performances but with 7$\times$ less steps. Notably, Peak Signal-to-Noise Ratio (PSNR) improves from 28.02 to 45.33 and MAE from 0.324 to 0.059 in just 159$s$ and 20$\times$ fewer steps than the DAO ones.
Importantly, unlike RL-based planners dependent on large RT datasets, \textsc{L2Plan} is trained with only 180 outer-loop meta-optimization steps.
This underscores how learning update dynamics enable more effective escape from saddle points and faster convergence in training and inference. 
Furthermore, it is noteworthy that the generative fluence map deliverable manifold fidelity influences the accuracy of the dose matching. Indeed, despite the inherent differences between Frechet Inception Distance (FID) and Mean Squared Error (MSE) evaluation, our \textsc{FMD} one-shot generator achieves superior dose consistency than VAE-, GAN- and one-shot diffusion-based alternatives surpassing the leading diffusion model solution both in FID, Structural Similarity Index Measure (SSIM) by 2.2\% (0.963 \textit{v.s.} 0.985) and in PSNR by 1.41  (38.06 \textit{v.s.} 39.47) at 300 steps Adam convergence. Remarkably, while alternative optimizers are constrained by stable initialization of optimization parameters, \textsc{L2Plan} operates without it. Nevertheless, our method also outperforms the most recent L2O strategies by 2.1\% in SSIM (0.993 \textit{v.s.} 0.972), 4.71 in PSNR (45.33 \textit{v.s.} 40.54) and 3.2\% in MAE (0.059 \textit{v.s.} 0.091) being statistically different.

\begin{table}[t]
\centering
\caption{Effectiveness of each module in our method. The best results are in bold, while the (*) and ($\dagger$) indicate 0.001 and 0.05 p-value significant differences.}
\label{tab:ablation}

\begin{minipage}[t]{0.49\textwidth}
\centering

\begin{adjustbox}{width=0.62\textwidth}
\begin{tabular}{@{}cccccccc@{}}
\toprule
\multicolumn{5}{c}{\textbf{\textsc{FMD} Loss}} & &
\multicolumn{2}{c}{\textbf{Metric}} \\
\midrule
$\mathcal{L}_{\text{EDM}}$ & &
$\mathcal{L}_{\text{DMD}}$ & &
$\mathcal{L}_{\text{GAN}}$ & &
\textbf{Steps}$\downarrow$ & 
\textbf{FID}$\downarrow$ \\
\cmidrule(r){1-1} \cmidrule(lr){3-3}
\cmidrule(lr){5-5} \cmidrule(l){7-8}

\checkmark & & $\times$ & & $\times$ & & 18 & 15.40 \\
\checkmark & & \checkmark & & $\times$ & & 1 & 28.18 \\
\checkmark & & $\times$ & & \checkmark & & 1 & 22.76 \\
\checkmark & & \checkmark & & \checkmark & & \textbf{1} & \textbf{17.52} \\

\bottomrule
\end{tabular}
\end{adjustbox}
\end{minipage}
\hfill
\begin{minipage}[t]{0.50\textwidth}
\centering
\begin{adjustbox}{width=\textwidth}
\begin{tabular}{@{}clclclccc@{}}
\toprule
\multicolumn{6}{c}{\textbf{\textsc{L2Plan} Loss}} 
& \multicolumn{3}{c}{\textbf{Metric}}                                                                                                                \\ 
\cmidrule(r){1-6} \cmidrule(l){7-9} 
$\mathcal{L}_{\text{D}}$  &  & 
$\mathcal{L}_{\text{cont}}^{\mathrm{z}}$ &  & 
$\mathcal{L}_{\text{cont}}^{\mathbf{MU}}$ 
&  & 
\textbf{PSNR} $\uparrow$ & 
\textbf{MAE$_{PTV}$}$\downarrow$ & 
\textbf{MAE$_{OARs}$}$\downarrow$ \\ 
\cmidrule(r){1-1} \cmidrule(lr){3-3} \cmidrule(lr){5-5} \cmidrule(l){7-9} 

\checkmark &  & $\times$     &  & $\times$     &  & 43.22$\pm$1.51* & 0.509$\pm$0.197* & 0.168$\pm$0.086* \\
\checkmark &  & \checkmark   &  & $\times$     &  & 44.76$\pm$1.47$\dagger$ & 0.469$\pm$0.198$\dagger$ & 0.163$\pm$0.086$\dagger$ \\
\checkmark &  & $\times$     &  & \checkmark   &  & 44.56$\pm$1.01$\dagger$ & 0.480$\pm$0.178$\dagger$ & 0.162$\pm$0.016$\dagger$ \\
\checkmark &  & \checkmark   &  & \checkmark   &  & \textbf{45.33$\pm$1.62} & \textbf{0.445$\pm$0.108} & \textbf{0.154$\pm$0.087} \\ 

\bottomrule
\end{tabular}
\end{adjustbox}

\end{minipage}

\end{table}

\subsection{Ablation Study} 
\Cref{tab:ablation} reports ablations of the \textsc{FMD} and \textsc{L2Plan} losses. Using EDM as a teacher, reducing denoising steps slightly increases FID. However, $\mathcal{L}_{\text{DMD}}$ and $\mathcal{L}_{\text{GAN}}$ act complementarily: the former preserves alignment with the teacher, while the latter corresponds alone to the D2O model~\cite{zheng2025revisiting}, enhances perceptual realism. Their combination achieves the best trade-off in manifold fidelity (17.52 \textit{v.s.} 15.40 FID) and efficiency (1 \textit{v.s.} 18 steps). On the optimization side, introducing continuity priors to $\mathcal{L}_{\text{D}}$ progressively improves planning quality. $\mathcal{L}_{\text{cont}}^{\mathrm{z}}$ increases PSNR by 1.54 and reduces MAE$_{PTV}$, while $\mathcal{L}_{\text{cont}}^{\mathbf{MU}}$ further stabilizes OARs errors. Combining both priors yields the best overall performance, demonstrating that regularization enhances dose consistency and stability during refinement.

\begin{figure}[t]
\centering
\includegraphics[width=\textwidth]{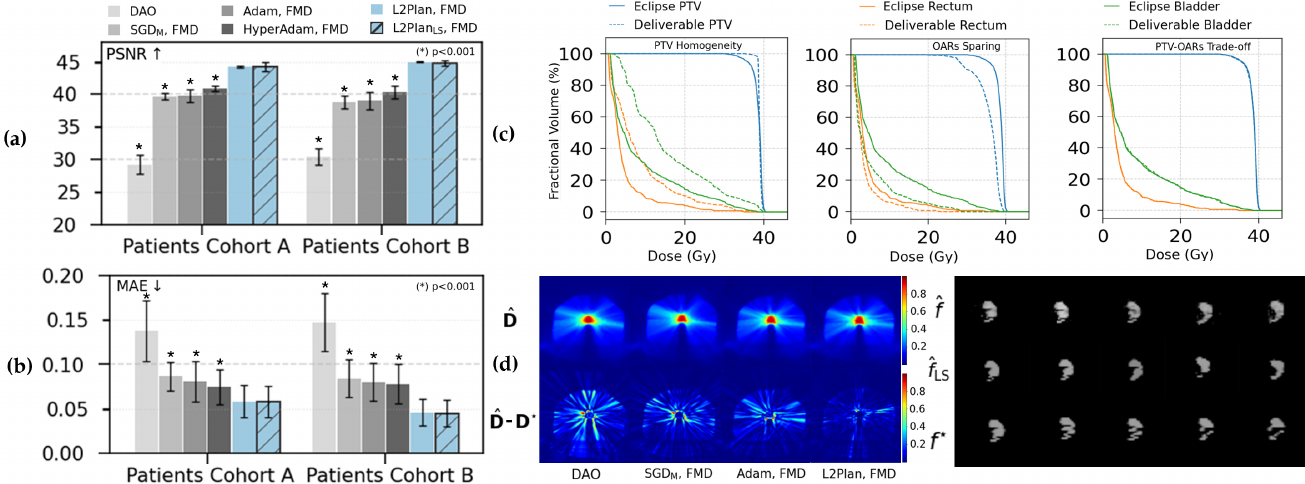}
\caption{Results on two private patients cohorts (a, b), flexibility DVH analysis after LS~(c), visual results of \textsc{L2Plan} dose ($\mathbf{\hat D}$), the comparison with its target ($\mathbf{D^*}$) and a set of contiguous CPs of \textsc{L2Plan}-, LS- and target fluence maps ($\hat f$, $\hat f_\text{{LS}}$, $f^*$)(d).} \label{fig:vis_res}
\end{figure}
\subsection{Results with Other Datasets}
Since \textsc{L2Plan} operates as an optimizer at inference, it is expected to generalize its update dynamics to any patient‑cohort and RT‑plan objective without retraining, provided the task-loss share similar structure. We evaluate \textsc{L2Plan} on two private prostate cohorts (135 and 13 patients) from different institutions with objectives differing from REQUITE (\Cref{fig:vis_res}~(a,b)). Across both cohorts, our approach improves PSNR and MAE by up to 4.01 and 0.7, respectively, over standard RT planners. It also surpasses the best L2O baseline by 3.41 and 4.65 PSNR and 0.016 and 0.032 MAE on cohorts A-B, demonstrating generalizability. 

\subsection{Results on Clinically Oriented Application Tasks}
We further evaluated \textsc{L2Plan} for plan consistency and flexibility essential for RT practice translation. Using a rule‑based LS tool, we confirm that our optimized fluence maps satisfy Varian M120 MLC machine‑mechanical constraints, with statistically significant pre‑/post‑LS agreement across REQUITE (\Cref{tab::res}) and private cohorts (\Cref{fig:vis_res}~(a,b,d)). For flexibility, we add a $\mathcal{L}_1$ term to $\mathcal{L}_{\text{plan}}$ to promote 38$Gy$ PTV homogeneity or OAR sparing. We show that the model accommodates these objectives without retraining in \Cref{fig:vis_res}~(c): enforcing PTV homogeneity achieves a Homogeneity Index (HI) of 0.041, with expected OAR sparing reduction, whereas prioritizing OAR protection reduces their minimum dose to 0.7$Gy$, with PTV HI naturally decreasing to 0.233. \emph{Focusing on clinical application}, these results show how \textsc{L2Plan} can \emph{adapt consistently and on‑the‑fly to various objectives, directly supporting RT physicians in practice.}

%% file: sections/conclusions.tex
We present a diffusion L2O method for end-to-end VMAT planning. A distilled diffusion model learns a clinically feasible fluence manifold, while an LSTM-based optimizer captures adaptive gradient dynamics to satisfy prescribed dose objectives efficiently. Unlike most recent RT-planners that suffer from plateau-dominated landscapes, rely on tuned initialization, and compute-intensive re-optimization when objectives change, our validation on public and private data-sets demonstrates initialization-agnostic, saddle-point-free, flexible, and machine-consistent planning, bringing VMAT planning one step closer to more practical deployment. Future work will explore one-shot latent diffusion strategies to reduce $\bm{\phi}$ number of optimized parameters and further enhance VMAT planning efficiency.